\documentclass[
]{ceurart}

\usepackage{listings}
\usepackage{nicematrix,booktabs,subcaption}
\begin{document}

\copyrightyear{2025}
\copyrightclause{Copyright for this paper by its authors.
  Use permitted under Creative Commons License Attribution 4.0
  International (CC BY 4.0).}

\conference{TRUST-AI: The European Workshop on Trustworthy AI. Organized as part of the European Conference of Artificial Intelligence - ECAI 2025. October 2025, Bologna, Italy.}

\title{Cross-Layer Attention Probing for Fine-Grained Hallucination Detection}

\author[1]{Malavika Suresh}[%
email=m.suresh@rgu.ac.uk,
]
\cormark[1]
\address[1]{Robert Gordon University, Aberdeen, United Kingdom}

\author[2]{Rahaf Aljundi}[%
email=rahaf.al.jundi@toyota-europe.com,
]
\address[2]{Toyota Motor Europe, Brussels, Belgium}

\author[1]{Ikechukwu Nkisi-Orji}[%
email=i.nkisi-orji@rgu.ac.uk,
]

\author[1]{Nirmalie Wiratunga}[%
email=n.wiratunga@rgu.ac.uk,
]

\cortext[1]{Corresponding author.}

\begin{abstract}
With the large-scale adoption of Large Language Models (LLMs) in various applications, there is a growing reliability concern due to their tendency to generate inaccurate text, i.e. \textit{hallucinations}.
In this work, we propose Cross-Layer Attention Probing (CLAP), a novel activation probing technique for hallucination detection, 
which processes the LLM activations across the entire residual stream as a joint sequence.
Our empirical evaluations using five LLMs and three tasks show that CLAP improves hallucination detection compared to baselines on both greedy decoded responses as well as responses sampled at higher temperatures, thus enabling fine-grained detection, i.e. the ability to disambiguate hallucinations and non-hallucinations among different sampled responses to a given prompt. This allows us to propose a detect-then-mitigate strategy using CLAP to reduce hallucinations and improve LLM reliability compared to direct mitigation approaches.
Finally, we show that CLAP maintains high reliability even when applied out-of-distribution. Code available at: \url{https://github.com/itsmemala/CLAP}
\end{abstract}


\maketitle

\section{Introduction}
Large Language Models (LLMs) have become increasingly accessible and scalable for commercial use, largely due to the API-based access offered by several LLM platform providers. From AI-generated summaries in search engines to chatbots in various health and business sector applications, LLM generated text is being widely consumed by a large proportion of the population. Such widespread adoption increases the risk of spreading misinformation and causing harm to users through factually incorrect LLM-generated text, i.e. \textit{hallucinations}. Improving the trustworthiness by detecting and mitigating hallucinations is therefore an important research objective.

Current approaches for tackling LLM hallucinations fall under three broad categories: black-box, grey-box and open-box methods. 
Among open-box methods, some recent works \cite{kossen2024SEProbes,azaria2023SAPLMA,burns2022ccs} have established the potential of building binary hallucination detectors using raw LLM activations as input, termed \textit{activation probing}. Other works have shown that hallucinations can be mitigated by directly editing activations \cite{chuang2023dola,li2023inferencetimeintervention} or output probabilities \cite{Huang2023OPERA} at generation time. However, the effect of activation-editing on other model capabilities is not well understood. While prior work on activation probing has focused on individual layers, in this work we introduce Cross-Layer Attention Probing (CLAP), a fundamentally different approach that utilises the full residual stream (i.e. activations from all LLM layers) to probe model behaviour more comprehensively. We hypothesise that the contribution of activations at different layers to hallucination detection varies for different tasks. To extract the most relevant information across layers, our method constructs a sequence of tokens by considering the activations at each LLM layer as an input token, and employs an attention mechanism over the sequence input. The design of our proposed probing technique is motivated by prior studies investigating the role of different LLM layers in language generation \cite{schuster2022earlyexit,geva2022ffnlayers} and hallucinations \cite{chuang2023dola}. We model hallucination detection as a supervised classification problem, which is supported by recent work \cite{karbasi2025impossibilityautomatedhallucinationdetection} that shows that automatic hallucination detection is not possible without both positive and negative examples. First, across five LLMs and three tasks (two factual question-answering tasks and one chain-of-thought reasoning task), we show that our method improves over uncertainty baselines and activation probing methods that consider only individual layers. 

Next, we build on the observation that different responses sampled for a given prompt can vary, with some being hallucinated and others not. We leverage the responses in the sampled space to augment the training data. 
We find that our proposed method, by learning to attend to different layers, can leverage this fine-grained supervision signal better to provide improved fine-grained detection performance compared to baselines. We further explore the integration of our method with hallucination mitigation pipelines, such as DoLa~\cite{chuang2023dola}. Noting that mitigation methods can adversely affect originally non-hallucinated samples, we combine CLAP with DoLa. Our results demonstrate that this combination significantly reduces the hallucination rate.

To support our approach of attending to activations across layers, we rigorously evaluate our method against various strategies for selecting probes at different layers. We conduct tests using cases from domains different from the training data to assess the generalization of layer-based probes compared to our proposed solution. Our results show that CLAP provides significant gains over probing at different layers when prompts fall outside the domains of the training samples. Notably, CLAP also improves over Semantic Entropy Probes \cite{kossen2024SEProbes}, which have been shown to generalise well.

In summary, this paper makes the following contributions:
\begin{enumerate}
    \item A novel probing technique, called Cross-Layer Attention Probing (CLAP), which consists of an attention mechanism operating on the LLM residual stream, is proposed for improving hallucination detection.
    \item CLAP improves fine-grained detection of hallucinations among different responses sampled for the same prompt, helping reduce model hallucinations.
    \item On an out-of-distribution study, CLAP improves over probes constructed at individual layers.
\end{enumerate}

The rest of the paper is structured as follows. Section \ref{related-works} describes methods used in prior work for hallucination detection and mitigation. Section \ref{method} describes our proposed approach, CLAP, and the methodology for fine-grained detection and mitigation. Section \ref{experiments} evaluates CLAP against baselines. Section \ref{analysis} provides an analysis of out-of-distribution generalisation. Finally, we perform an ablation study of the design in section \ref{ablation} before concluding with a discussion on future work in section \ref{conclusion}.

\section{Related work}\label{related-works}

\subsection{LLM-based detection and mitigation (black box)} Black-box methods assume no access to model internals and therefore rely on additional LLM-prompting. \cite{manakul2023selfcheckgpt} proposed the use of a \textit{consistency} check among different responses sampled for a given prompt as a measure of hallucination. Here the assumption is that when an LLM hallucinates, the sampled responses would be inconsistent with each other. This evaluates whether for a given prompt, \textit{any given response} from the LLM can be trusted, and is therefore not suited to identify non-hallucinating responses within the sampled space for a prompt as well as in cases where multiple different answers are valid. Other works \cite{mundler2023selfcontra,dhuliawala2023cove} have shown that LLMs can be prompted to detect hallucinations in outputs by the same or different LLM. This relies on the LLM having a good reasoning ability and is therefore often restricted to large models, introducing additional cost and latency.

\subsection{Uncertainty estimation (grey-box)} Uncertainty estimation methods \cite{manakul2023selfcheckgpt,Kuhn2023SemanticUncertainty,duan2023tokensar} use the probabilities of the generated output tokens to measure the confidence or \textit{uncertainty} in the generation, using a threshold  to classify low confidence outputs as hallucinations. However, identifying an appropriate threshold is often challenging, especially for long output sequences. Instead of considering the uncertainty in a single LLM generated response, \cite{Kuhn2023SemanticUncertainty} propose to measure the \textit{semantic uncertainty} in a set of responses sampled from the LLM for a given prompt. The authors show that a high semantic uncertainty (i.e. high conceptual variety) in sampled responses is a good indication of hallucination. The confidence estimate in this case is similar to the \textit{consistency} measure and has the same pitfalls mentioned above. Overall, current uncertainty estimates, by relying purely on the output probabilities, remain naive approaches to hallucination detection.

\subsection{Activation probing (open-box)} Recent works \cite{burns2022ccs,azaria2023SAPLMA,li2023inferencetimeintervention,kossen2024SEProbes} have focussed on building hallucination detectors or \textit{probes} using the LLM activations at generation time. While ITI \cite{li2023inferencetimeintervention} constructs the probes using activations at the output of attention heads, CCS \cite{burns2022ccs}, SAPLMA \cite{azaria2023SAPLMA}, Semantic Entropy probing \cite{kossen2024SEProbes} and HaloScope \cite{du2024haloscope} use the activations at the output of the transformer decoder block (layer activations). Unlike ITI and SAPLMA, where probes are trained in a supervised manner using a dataset of labelled responses, HaloScope and CCS train the probes in an unsupervised manner. 
Some works provide interpretability -
\cite{chuang2024lookbacklens} show that hallucinations with respect to input context are caused by the LLM attending to generated tokens rather than context tokens, while \cite{yuksekgonul2024attentioncsp} show that lack of attention to entity tokens is indicative of lack of entity knowledge. \cite{ferrando2025sae} show that there exist latent directions in the layer activation space that correspond to notions of "I know this entity" and "I don't know this entity". Unlike these prior works that focus on activations at specific points/layers during decoding, in this paper we propose an approach to improve detection by extracting a signature of hallucination across the entire residual stream.

\subsection{Activation editing (open-box)} Several studies aim to mitigate hallucinations during the decoding process by manipulating model activations \cite{li2023inferencetimeintervention}, adjusting token output probabilities \cite{shi2023CAD,chuang2023dola} or modifying output logits \cite{Huang2023OPERA}. In ITI \cite{li2023inferencetimeintervention}, model activations are shifted towards a direction associated with `truthfulness'. CAD \cite{shi2023CAD} contrasts output token probabilities generated with and without the input context to obtain new token probabilities that are expected to be more aligned with the input context. DoLa \cite{chuang2023dola} builds on the early exit strategy \cite{schuster2022earlyexit} and contrasts the output probabilities of the final layer with those of the intermediate layers. In Opera  \cite{Huang2023OPERA}, the final layer logits are modified with a penalty term that discourages the model from attending to summary tokens in long-form generation tasks.

\begin{figure*}[t!]
    \centering
    \includegraphics[width=1\textwidth]{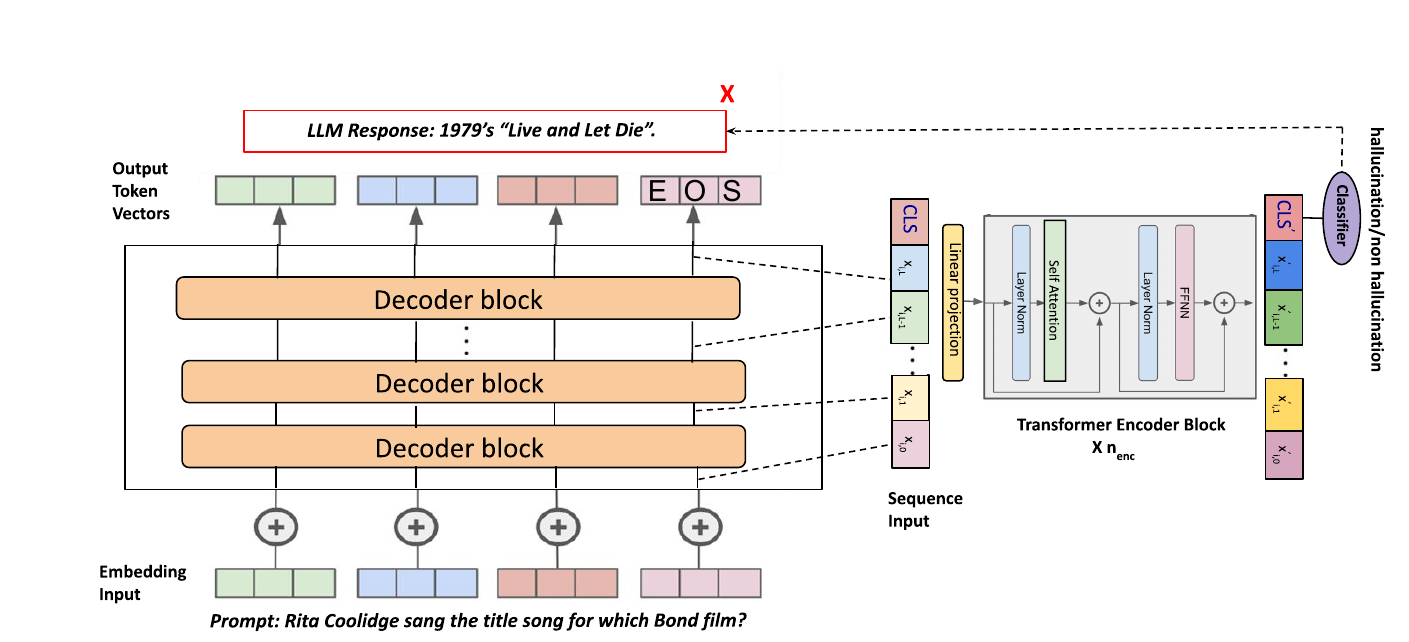}
    \caption{Cross-Layer Attention Probing: For an LLM with $L$ layers (transformer decoder blocks) and an input prompt $i$, activations $x_{i,l}$ are collected at the output of each layer $l$ when generating the last response token (EOS). The activations are first projected to a lower dimensional space through a learnable projection layer and the set of all projected activations forms a sequence of input tokens. A learnable CLS token is employed at the start of the sequence to extract features of hallucination. The sequence input is then fed to a transformer encoder block and the CLS embedding output is fed to a linear classifier layer for binary label prediction of hallucination/non hallucination decision.}
  \label{fig:method2}
\end{figure*}

\section{Method}\label{method}

In this section, we propose a novel probing technique that incorporates a learning mechanism over the activations at different layers. Several works have indicated that due to the residual connections in transformers, the outputs of individual LLM layers can be considered to be in the same embedding space \cite{schuster2022earlyexit,geva2022ffnlayers}. Building on this stream of work, here we explore how the activation pattern across LLM layers can be exploited to better detect hallucinations as compared to looking at activation patterns of individual layers. Unlike \cite{chuang2023dola}, where the authors propose to contrast the output of the final LLM layer against that of intermediate layers as an alternative decoding strategy, here we leverage the residual stream for improving hallucination detection. To this end, we propose \textit{cross-layer attention probing (CLAP)}, which takes as input the activations across all LLM layers when generating a given token.

\paragraph{Notations}\label{method-strategies}
Consider a dataset $D = \{P,R\}$ of prompts, $P$ and corresponding LLM responses, $R$. For a given prompt $p_i \in P$ passing through an LLM with $L$ layers, let $x_{i,l} \in \mathbb{R}^{d_{LLM}}$ represent the activation vector at layer $l$ of the LLM, where $d_{LLM}$ is the LLM activation dimensions. Following prior work \cite{li2023inferencetimeintervention,azaria2023SAPLMA,burns2022ccs}, we probe the activations when generating the last token of the LLM output response (EOS token). Let $y_i$ represent the binary label of hallucination/non-hallucination for the corresponding LLM response $r_i \in R$. We assume that ground truth correct answers to prompts are available and compare the model generated response to ground truth to obtain this label.

\subsection{Cross-Layer Attention Probing \textbf{(CLAP)}}\label{method-transformer}

Figure \ref{fig:method2} depicts our proposed probing method. First, we consider the set of all layer level activations $\{x_{i,l}\}$ as forming a set of input tokens. The tokens are arranged in the same order as the LLM layers (i.e. the residual stream) in order to be processed jointly as a sequence input. Depending on the dimensions of the LLM being probed, the sequence input can get very large, increasing computational costs. In order to allow scaling the method to larger LLMs, the activations are passed through a learnable down-projection layer at the start to produce $x'_{i,l} \in \mathbb{R}^{d_{model}}$.

The down-projected sequence input is then fed through a transformer encoder block, with $n_{enc}$ encoder layers (we experiment with $n_{enc} \in \{1,2\}$), each consisting of a self-attention module and a feed-forward network. The role of this encoder block is to learn to extract a pattern of hallucination across the residual stream by attending differently to activations of different layers and thus learn an embedding vector that better separates hallucinating and non-hallucinating responses. To extract this information, we employ a learnable CLS token at the start of the sequence input. This transforms the setting into a supervised classification problem, and the transformer embedding output at the CLS position is then fed to a linear classifier layer and trained with binary cross-entropy using the supervision signal $y_i$. 

\subsection{Leveraging Hallucinations in the Sampled Space for Fine-Grained Detection} The sampled response space for a given prompt can contain both hallucinations and non-hallucinations, indicating that correct entity/information can in fact exist in the residual stream even when the most confident generation is incorrect \cite{chuang2023dola}. Given that our proposed probing mechanism attends to the activations across the entire residual stream, we hypothesise that it can also be applied for a fine-grained detection of hallucination among responses sampled for the same prompt. In order to guide the probe training for fine-grained detection, we sample a set of $K$ additional responses to each prompt at high temperature, alongside the greedy decoded response. Each response is then labelled independently as hallucination/non-hallucination. When including the sampled responses during training all responses generated for a given prompt are always arranged in the same batch - we ablate this choice against random sampling in appendix \ref{ap:batching-sampled-responses}. We use CLAP trained on the set of all greedy and sampled responses to prompts as the method for detecting hallucinations at the sample level, making it compatible with different strategies of decoding/sampling responses.

\subsection{Hallucination Mitigation} 
\label{method-mitigation}
Strategies that aim to mitigate hallucinations by directly modifying activations or output token probabilities during decoding can negatively impact the quality of original, non-hallucinated responses, as we shall demonstrate in our experiments in section \ref{results}. A natural approach to address this issue is to couple the hallucination mitigation strategy with hallucination detection. In this section, we discuss how CLAP can be employed for this purpose. Given a fine-grained CLAP hallucination detector trained for a given LLM, we use the macro-F1 score on an in-distribution validation set to determine a classification threshold for binary hallucination label prediction. Then at test time, we generate responses with CLAP as follows:
\begin{enumerate}
    \item Generate greedy decoded response.
    \item Classify whether the response is hallucinated using CLAP.
    \item When classified as hallucination, generate an alternative response using either DoLa decoding \cite{chuang2023dola} or random sampling.
    \item Classify whether the alternate response is hallucinated using CLAP.
    \item Abstain when both the greedy response and alternate response are classified as hallucination.
\end{enumerate}

In summary, we combine default decoding with an alternate response on a \textit{per need basis} to improve hallucination mitigation, without the negative effects of directly applying mitigation strategies such as DoLa. When the mitigation strategy is signalled to fail by CLAP we abstain from responding, leading to safer use of LLMs.

\section{Experiments}\label{experiments}
Section \ref{setup} describes the setup used for the main experiments.
Section \ref{results} presents the results.
\subsection{Experimental setup}\label{setup}

\paragraph{Data} Experiments are conducted on two open-domain question answering (QA) tasks - Natural Questions (\textbf{NQ}) \cite{lee2019naturalquestions} and Trivia QA (\textbf{TQA}) \cite{joshi2017triviaqa} - and one chain-of-thought (COT) reasoning task - Strategy QA (\textbf{STR}) \cite{geva2021strategyqa}. The LLMs are evaluated in a closed-book setting for each of the tasks. Prompt formats used are shown in appendix \ref{ap:prompt-formats}. For each prompt, greedy decoding is used to generate the response. 
When generating additional sampled responses per prompt, sampling temperature and top\_p parameter are set to 1 and 0.95, respectively. See appendix \ref{ap:dataset-stats} for notes on data labelling and dataset statistics. In appendix \ref{ap:resp-refusal}, we ablate the rate of true hallucinations versus query refusals.

\paragraph{Models}  We use Llama-7B \cite{touvron2023llama7B}, Alpaca-7B \cite{alpaca}, Vicuna-7B \cite{vicuna2023}, Gemma-2B \cite{gemmateam2024gemma} and Llama3.1-Instruct-8B \cite{grattafiori2024llama3herdmodels} in our experiments.

\paragraph{Implementation Details} For CLAP, we set the linear projection dimension $d\_model = 128$  and use a held-out validation set to select the number of encoder layers $n_{enc} \in$ [1,2] keeping the memory footprint low. We report results of varying $d\_model$ in section \ref{ablation}. Further details are in appendix \ref{ap:imp-details}.

\paragraph{Baselines} Our main focus is in comparing the accuracy of probes that consider only the final layer activations to that of probing techniques that consider multiple layers. Therefore, for our main baselines, we report the results of a (1) linear probe \textbf{LP} and a (2) non-linear probe \textbf{NLP} \cite{azaria2023SAPLMA} on the last layer activations. Additionally, for reference, we report the results of a (3) classifier based on the predictive entropy \textbf{PE} \cite{Kuhn2023SemanticUncertainty} of the generated text and a (4) linear probe on the attention head activations \textbf{AH} \cite{li2023inferencetimeintervention} - for which we use the best performing head identified using a held-out validation set.

\subsection{Results}\label{results}

\paragraph{CLAP for fine-grained hallucination detection}
Table \ref{tab:main-average} compares the hallucination detection performance of CLAP against the uncertainty and activation probing baselines on greedy and sampled responses to prompts. Dataset-wise expanded results are provided in appendix \ref{ap:main-results-expanded}. When testing on greedy responses, CLAP trained on greedy responses (CLAP-g)
generally improves over the baselines (PE, AH-g, LP-g, NLP-g), while including sampled responses at
train-time can often provide further gains for CLAP (CLAP-s). AH performs slightly better than CLAP
on Gemma-2B and PE performs slightly better than CLAP on Lamma3.1-Instruct-8B. However these
baselines are inferior to CLAP when coupled with other LLMs.
When testing on sampled responses, we find that CLAP can leverage the sampled responses at train-time (CLAP-s) better than the baselines (AH-s, LP-s, NLP-s) to improve fine-grained detection consistently and providing gains of up to 1.5\% (on TQA with Alpaca-7B and Gemma-2B). Though AH-s performs slightly better than CLAP on average with Gemma-2B, CLAP couples more robustly with all the LLMs, illustrating that it is agnostic to the LLM and widely applicable.

\begin{table*}[h]
\centering
\caption{Hallucination detection performance of CLAP versus baselines, measured in AUC
scores, best is indicated in bold, second best is underlined. '-g' and '-s' denote the use of greedy responses only and the use of both greedy and sampled responses for training, respectively. Results averaged across TQA, NQ and STR, across three random seeds.}\label{tab:main-average}
\begin{tabular}{@{}crrrrrrrrr@{}}
\toprule
LLM &PE &AH-g &LP-g &NLP-g &CLAP-g &AH-s &LP-s &NLP-s &CLAP-s \\
\midrule
\multicolumn{10}{c}{Greedy Test Responses}\\
\midrule
Llama 7B &69.0 &64.0 &77.7 &77.2 &\textbf{78.1} &69.2 &75.3 &75.6 &\underline{77.8}\\
Alpaca 7B &78.6 &81.2 &85.2 &85.8 &\underline{86.6} &83.6 &86.3 &86.4 &\textbf{87.3}\\
Vicuna 7B &80.4 &78.0 &88.4 &88.2 &88.5 &85.7 &\textbf{88.8} &\textbf{88.8} &\underline{88.7}\\
Gemma 2B &62.8 &\underline{73.5} &70.2 &71.2 &72.7 &\textbf{74.1} &70.1 &71.6 &72.7 \\
Llama3.1-I 8B &\textbf{69.7} &66.6 &66.1 &66.5 &68.1 &67.6 &67.8 &68.3 &\underline{69.0}\\
\midrule
\multicolumn{10}{c}{Sampled Test Responses}\\
\midrule
Llama 7B &84.3 &78.2 &75.8 &77.5 &74.3 &86.5 &88.8 &\underline{89.0} &\textbf{89.9} \\
Alpaca 7B &79.7 &81.5 &84.8 &84.4 &84.8 &83.8 &86.0 &\underline{86.8} &\textbf{88.1} \\
Vicuna 7B &87.9 &87.0 &90.8 &90.9 &91.1 &90.8 &\underline{93.1} &\textbf{93.3} &\textbf{93.3}  \\
Gemma 2B &72.4 &69.3 &65.4 &66.2 &69.7 &\textbf{76.8} &73.4 &74.5 &\underline{76.5} \\
Llama3.1-I 8B &73.8 &67.6 &68.3 &68.0 &69.6 &69.8 &73.5 &\underline{74.1} &\textbf{74.9}\\
\bottomrule
\end{tabular}
\end{table*}

\paragraph{Improving hallucination mitigation with CLAP}
In this section, we show how fine-grained detection using CLAP can help improve hallucination mitigation. 
Table \ref{tab:mitigation} compares the percentage of non-hallucinated responses using our approach of combining CLAP with mitigation (denoted +CLAP-II), as described in section \ref{method-mitigation}, alongside four baseline strategies, described below:

\begin{itemize}
    \item \textbf{Default (Def)} Always use the greedy decoding strategy. 
    \item \textbf{Def+Abstain} 1. Generate greedy decoded response. 2. Classify whether hallucinated using CLAP. 3. Abstain when classified as hallucination.
    \item \textbf{Alternate (Alt)} Always use an alternate, non-greedy decoded response. Here we use DoLa \cite{chuang2023dola}.
    \item \textbf{+CLAP-I} 1. Generate greedy decoded response. 2. Classify whether hallucinated using CLAP. 3. When classified as hallucination, generate an alternate response.
\end{itemize}

First, we see that with the +CLAP-I strategy, non-hallucination rate is generally improved over the Default and Alternate strategies, with an overall average gain of \textbf{11.7\%} over Default and \textbf{4.7\%} over Alt. Next, with the +CLAP-II strategy, we additionally detect hallucinations in the alternate response and abstain if hallucinated. We see that +CLAP-II reduces the abstention rate significantly (by \textbf{24.5\%} on average) compared to the Def+Abs strategy while consistently maintaining high non-hallucination rate among non-abstained responses. Both these observations demonstrate the practical utility of our fine-grained CLAP detector for improving LLM reliability. In appendix \ref{ap:mitigation-random-sampling}, we show similar gains when using random sampling as the alternate response.

\begin{table*}[h]
\centering
\caption{Mitigating hallucinations with fine-grained detection using CLAP applied to DoLa. 
First block indicates the LLM and dataset. Second block shows the \% of non-hallucinated responses for each of the five response generation strategies. Third and forth blocks show the \% of responses abstained and the \% of responses abstained but were non-hallucinated (NH), respectively. For results with +CLAP-I and +CLAP-II we report mean across three seeds. * denotes \%NH among non-abstained responses.}\label{tab:mitigation}
\begin{NiceTabular}{@{}c|rrrrr|rr|rr@{}}
\toprule
&\multicolumn{5}{c}{\% Non-Hallucinations (NH) $\uparrow$} &\multicolumn{2}{c}{\%Abs $\downarrow$} &\multicolumn{2}{c}{\%Abs but NH $\downarrow$}\\
Data &Def &Def+Abs\textsuperscript{*} &Alt &+CLAP-I &+CLAP-II\textsuperscript{*} &Def+Abs &+CLAP-II &Def+Abs &+CLAP-II\\
\midrule
L-7B TQA &57.3 &75.6 &48.6 &56.3 &71.3 &36.0 &26.4 &9.1 &5.5\\
L-7B STR &60.0 &69.9 &64.0 &65.9 &67.6 &45.0 &7.2 &21.4 &3.9\\
\midrule
A-7B TQA &34.2 &68.3 &33.8 &41.9 &68.1 &62.7 &44.2 &8.7 &6.3\\
A-7B STR &43.4 &67.0 &60.5 &64.8 &65.2 &44.3 &16.1 &6.1 &2.6\\
\midrule
V-7B TQA &14.6 &59.2 &28.3 &31.0 &57.7 &80.7 &59.9 &3.2 &2.6 \\
V-7B STR &42.8 &65.6 &59.4 &62.7 &64.0 &40.3 &8.3 &3.6 &0.7 \\
\midrule
Average &42.1 &67.6 &49.1 &53.8 &65.7 &51.5 &27.0 &8.7 &3.6\\
\bottomrule
\end{NiceTabular}
\end{table*}

\begin{figure*}[h!]
    \centering
    \begin{subfigure}{0.48\textwidth} 
        \centering
        \includegraphics[scale=0.3]{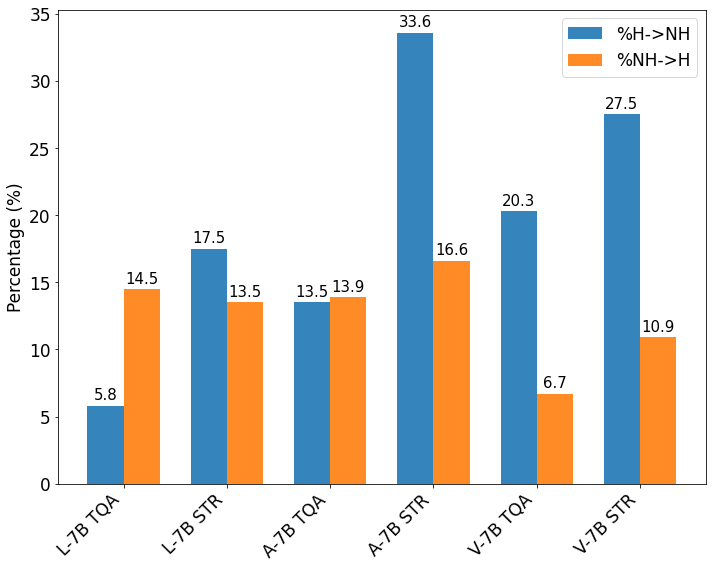}
        \caption{Shows the \% H->NH (or NH->H) transitions using Alt.}
        \label{fig:mitigation-changes-dola}
    \end{subfigure}
    \hfill
    \begin{subfigure}{0.48\textwidth}
        \centering
        \includegraphics[scale=0.3]{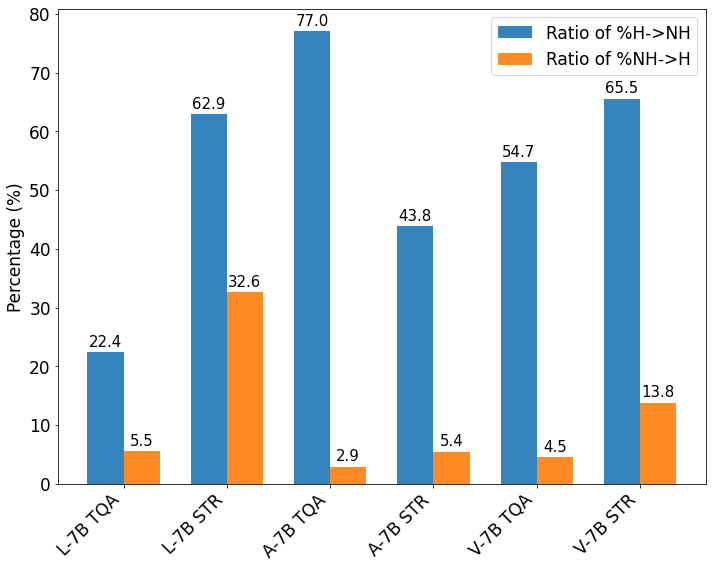}
        \caption{Shows the ratio of \% H->NH (or NH->H) transitions using CLAP-II compared to \% H->NH (or NH->H) transitions using Alt.}
        \label{fig:mitigation-changes-dola-vs-clapII}
    \end{subfigure}
    \caption{Transitions when replacing original greedy responses with responses using Alt and +CLAP-II strategies.}
    \label{fig:mitigation-changes}
\end{figure*}

In figure \ref{fig:mitigation-changes-dola}, we show the percentage of hallucinated greedy decoded responses that are replaced with non-hallucinated responses and vice versa when using the DoLa mitigation approach. We find that DoLa applied directly often negatively affects a significant percentage of the original non-hallucinated responses (orange bars). In figure \ref{fig:mitigation-changes-dola-vs-clapII}, we show the ratio of the replacement rate when using CLAP-II against the replacement rate when using DoLa directly. We see that CLAP-II significantly reduces the NH->H replacements (orange bars) while generally maintaining a good H->NH replacement rate (blue bars), thereby maximising the gains from DoLa.

In appendix \ref{ap:mitigation-clap-vs-baselines}, we show that mitigation using CLAP outperforms mitigation using baseline probes.

\section{Attending to layers benefits generalisability}\label{analysis}
In this section, we 
compare the out-of-distribution performance of CLAP to independent probes constructed at each LLM layer when transferring from one domain to another.
In addition to TQA and NQ, we use three categories from wikidata \cite{wikidata} (city-country, player-date-birth, movie-cast).
We construct twenty train-test pairs using these five datasets, which allows us to capture a wide array of generalisation scenarios.
At train time, for an LLM with $L$ layers, we construct $L$ independent probes, where each probe is a binary logistic regression classifier trained on the activations at one LLM layer $\{x_{i,l}\}$, to predict a hallucination (H)/non-hallucination (NH) label ${\{y_i\}}$. At test time, to classify an LLM response, we experiment with four strategies for selecting among the $L$ probe predictions, as follows.

\begin{itemize}
    \item \textbf{Last layer} Uses the probe trained on the last layer activations. 
    \item \textbf{Most Accurate Layer (MA)} Uses the in-distribution validation split to select one out of the $L$ probes that performs best for the domain trained on.
    \item \textbf{Most Confident Layer (MC)} Instead of pre-selecting a probe at train-time as above, this strategy measures the entropy of the predicted labels at each probe to then identify the probe with the most confident prediction (i.e., least entropy) for a given sample at test-time.
    \item \textbf{Majority Voting Across Layers (MV)} Uses an ensemble setup where the final label for a sample is given by the majority vote across all probes.
\end{itemize}

In table \ref{tab:main-entitywise}, we show the \% gain-over-baseline (AUC) achieved by CLAP over the probe selection strategies as well as semantic entropy probes \cite{kossen2024SEProbes}, which have been shown to generalise well. 
CLAP not only outperforms other hallucination detection strategies on in-distribution samples but also demonstrates generalisability to samples from domains not covered in the training set. This is a crucial property - if hallucination detection deteriorates out-of-domain, the LLM is left with no guard. 

\begin{table*}[!h]
\centering
\caption{Out-of-distribution (OOD) hallucination detection performance of CLAP versus probe selection strategies and semantic entropy probing (SEP), measured in \% gain on AUC. Results averaged across twenty OOD pairs derived from five datasets, 
across three random seeds, using greedy responses for train and test.}\label{tab:main-entitywise}
\begin{tabular}{@{}crrrrr@{}}
\toprule
LLM &\%gain-over-last	&\%gain-over-MA	&\%gain-over-MC	&\%gain-over-MV	&\%gain-over-SEP\\
\midrule
Llama 7B &4.9 &2.1 &4.9 &1.7 &45.2 \\
Alpaca 7B &0.9 &1.4 &1.1 &1.3 &18.0 \\
Vicuna 7B &5.1 &1.3 &-1.8 &5.2 &35.6 \\
Gemma 2B &6.3 &3.6 &1.4 &3.8 &25.1 \\
Llama3.1-I 8B &2.4 &-1.6 &-2.2 &5.9 &15.1\\
\bottomrule
\end{tabular}
\end{table*}

\section{Ablating design choices for CLAP}\label{ablation}
First, in table \ref{tab:ablation-hyp-llama-auc}, we assess the sensitivity of CLAP to the number of encoder layers used and input dimensionality reduction. For TQA and NQ, increasing the projection dimensionality ($d_{model}$) has negligible effect while adding another encoder layer ($n_{enc}=2$) can result in a slight gain. For STR, performance is sometimes improved with higher projection dimensionality. We note that directly using raw activations or projecting to high dimensions becomes prohibitively expensive for larger LLMs. In this regard, we interpret our results as indicating that discriminative information for detecting hallucinations is retained at lower dimensions, making the method viable for larger LLMs. We note that CLAP with $d_{model}=128$ and $n_{enc}=2$ has only 15K parameters for an LLM of 2B parameters.

\begin{table}[h]
\centering
\caption{Analysis of hyper-parameter choices, using AUC on validation set, best is indicated in bold. * denotes no projection. Results on Llama 7B, averaged across three random seeds, using greedy responses for train and test.}\label{tab:ablation-hyp-llama-auc}
\begin{tabular}{@{}rrrrr|rrrrr@{}}
\toprule
$n_{enc}$ &$d_{model}$ &TQA & NQ  & STR &$n_{enc}$ &$d_{model}$ &TQA & NQ  & STR\\
\midrule
1 &128 &84.3 (0.2) &86.1 (0.8) &64.6 (0.7 &2 &128 &84.3 (0.2) &86.0 (0.8) &64.5 (0.8)\\
1 &256 &84.3 (0.3) &86.1 (1.0) &64.2 (0.8) &2 &256 &84.3 (0.2) &86.2 (1.0) &64.7 (0.3) \\
1 &512 &84.2 (0.1) &86.0 (1.1) &64.8 (0.8) &2 &512 &84.3 (0.3) &86.1 (1.1) &65.1 (0.3) \\
1 &1024 &84.2 (0.4) &86.0 (0.7) &64.7 (0.9) &2 &1024 &84.3 (0.5) &\textbf{86.3 (0.7)} &65.2 (1.0) \\
1 &2048 &84.1 (0.4) &85.8 (1.0) &64.4 (0.5) &2 &2048 &\textbf{84.4 (0.4)} &86.0 (1.1) &65.7 (1.2)\\
1 &4096* &83.8 (0.1) &86.0 (0.9) &63.6 (0.6)\ &2 &4096* &83.9 (0.4) &86.0 (1.0) &\textbf{66.7 (1.6)}\\
\bottomrule
\end{tabular}
\end{table}

Next, we ablate the design of CLAP in table \ref{tab:ablation-design-llama-auc}, by comparing against two alternative probes that also take activations from all LLM layers as input but without any cross-layer attention mechanism. \textbf{Maxpool} denotes the element-wise max-pooling of all activations before training a linear classifier layer. \textbf{Project + Concat} denotes the use of a learnable down-projection layer on layer-wise activations followed by concatenation of the projected activations before training a linear classifier layer. We see that Maxpool, though computationally more efficient, performs much worse than Project + Concat. This indicates the benefit of modelling layer-wise activations jointly. As we increase the projection dimensions, the performance of Project + Concat sometimes improves but computation cost increases significantly. The benefit of performing cross-layer attention is evident in the out-of-distribution tests, where CLAP ($n_{enc}=2, d_{model}=128$, 1.1M params) provides significant gains at comparable costs over Project + Concat ($d_{model}=256$, 1M params). In appendix \ref{ap:clap-vs-att-pool}, we compare CLAP with token-wise attention-pooling \cite{ch-wang-etal-2024-attpool}, showing again the advantage of CLAP in out-of-distribution testing.

\begin{table}[h]
\centering
\caption{Ablating CLAP components and design, using AUC on test set, best is indicated in bold, second best is underlined. * denotes no projection, K denotes one thousand and M denotes one million. Results on Llama 7B, averaged across three random seeds, using greedy responses for train and test.}\label{tab:ablation-design-llama-auc}
\begin{tabular}{@{}crrrcccc@{}}
\toprule
Probe &$n_{enc}$ &$d_{model}$ &\# Params &\multicolumn{2}{c}{In Distribution} &\multicolumn{2}{c}{Out Of Distribution}\\
 & & & &TQA &NQ &TQA->City &NQ->City\\
\midrule
Maxpool &- &4096* &4K &77.4 (0.7) &82.9 (0.3) &50.9 (2.8) &51.8 (2.5) \\
Project + Concat &- &128 &528K &82.7 (0.1) &\underline{87.0 (0.3)} &56.7 (1.2) &54.1 (0.9) \\
Project + Concat &- &256 &1M &82.6 (0.1) &86.9 (0.1) &56.8 (1.5) &\underline{54.4 (1.8)}  \\
Project + Concat &- &512 &2.1M &\underline{82.8 (0.1)} &86.9 (0.3) &57.0 (0.1) &53.2 (1.1)  \\
Project + Concat &- &4096* &16.9M &\textbf{83.1 (0.3)} &\textbf{87.1 (0.2)} &\textbf{57.5 (0.6)} &53.9 (0.7)  \\
\midrule
CLAP &1 &128 &862K &81.8 (0.5) &85.8 (0.8) &54.4 (1.3) &52.8 (2.2)  \\
CLAP &2 &128 &1.1M &82.0 (0.0) &86.6 (0.3) &\underline{57.4 (1.0)} &\textbf{55.8 (2.6)}  \\
\bottomrule
\end{tabular}
\end{table}

\section{Conclusion}\label{conclusion}
This work proposed a novel probing technique for detecting hallucinations in LLMs, called Cross-Layer Attention Probing (CLAP), that takes the entire LLM residual stream as a sequence of input tokens, with an attention mechanism operating over the layer-wise activations. CLAP outperforms uncertainty baselines and probes that consider only individual layers. Further, leveraging responses in the sampled space at train time helps CLAP achieve fine-grained detection between hallucinated and non-hallucinated responses to the same prompt at test time. This allowed us to apply CLAP as a fine-grained detector to reduce LLM hallucination rate by sampling alternative responses to a given prompt and distinguishing hallucinated outputs from non-hallucinated ones. Finally, an out-of-distribution study 
revealed that attending to different layers enables CLAP to generalise more effectively.

We focus on small LLMs of 2B-8B where hallucination is more prominent, making detection crucial. Our ablation study indicates that hallucinations can still be detected after projecting to lower dimensions, providing evidence for scaling CLAP to larger LLMs - we leave this to future work. While CLAP takes input from all layers, we leave the investigation of the role of each layer within CLAP to future work.



\section*{Declaration on Generative AI}
  The author(s) have not employed any Generative AI tools.
  

\bibliography{references}

\newpage
\appendix

\section{Experiment Setup}
\subsection{Prompt Formats}\label{ap:prompt-formats}
\begin{figure*}[h!]
    \centering
    \includegraphics[width=0.4\textwidth]{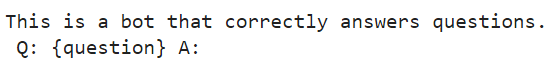}
    \caption{Prompt Format: Natural Questions, Trivia QA}
  \label{fig:prompt_format1}
\end{figure*}

\begin{figure*}[h!]
    \centering
    \includegraphics[width=0.55\textwidth]{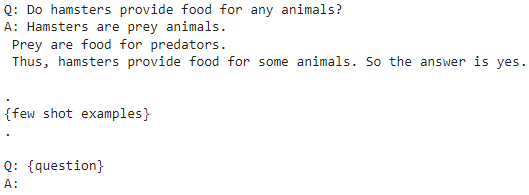}
    \caption{Prompt Format: Strategy QA}
  \label{fig:prompt_format2}
\end{figure*}

Figure \ref{fig:prompt_format1} \cite{Kuhn2023SemanticUncertainty} and figure \ref{fig:prompt_format2} \cite{chuang2023dola} show the prompt formats used for generating LLM responses for the three tasks considered in the main experiments.

\subsection{Data Labelling and Dataset Statistics}\label{ap:dataset-stats}
For Trivia QA and Natural Questions, each LLM response is labelled as hallucinated/non-hallucinated using a rouge-1 cut-off of 0.3, following prior work \cite{Kuhn2023SemanticUncertainty,duan2023tokensar}, where the rouge labels are validated against human annotated labels finding a 0.96 accuracy. For StrategyQA, each LLM response is labelled by matching the final answer produced after the COT against the gold reference of YES/NO, following prior work \cite{chuang2023dola}. Table \ref{tab:dataset-statistics} shows the number of prompts used, number of additional responses sampled per prompt at high temperature and the hallucination rate among greedy and sampled responses. We note that since we use LLMs of at most 8B parameters, given the wide range of facts queried and with no access to external information, such high hallucination rates are expected. For NQ with Llama3.1-I-8B, we find a very high hallucination rate of >95\% among greedy responses and exclude this from the analysis. We note that high hallucination rates for NQ are in line with observations in prior work \cite{wang2023TQA-NQ} and is generally attributed to the difference between typical LLM pre-training data and data used for creating NQ (Google search queries).

\begin{table}[h]
\centering
\caption{Statistics of the datasets used. Shows number of prompts, additional responses sampled per prompt and \% of hallucinated responses in the train and test splits.}\label{tab:dataset-statistics}
\begin{tabular}{@{}llrrrrrrrr@{}}
\toprule
& &\multicolumn{4}{c}{Greedy Responses} &\multicolumn{4}{c}{Sampled Responses}\\
\midrule
 Model &Dataset &\multicolumn{2}{c}{\# Prompts} &\multicolumn{2}{c}{\% Hallucinations} &\multicolumn{2}{c}{\# Samples} &\multicolumn{2}{c}{\% Hallucinations}\\
 & & Train & Test &Train &Test &Train &Test &Train &Test\\
\midrule
\multirow{3}{*}{Llama-7B} &TQA &5000 &1800 &45.9 &42.7 &10 &10 &79.1 &76.9\\
 &NQ &5000 &1800 &75.3 &77.4&10 &10 &93.5 &93.6\\
 &STR &1832 &458 &39.8 &40.0&8 &8 &68.3 &67.2\\
\midrule
 \multirow{3}{*}{Alpaca-7B} &TQA &5000 &1800 &67.5 &65.8 &10 &10 &74.6 &73.4\\
 &NQ &5000 &1800 &90.3 &89.9 &10 &10 &92.3 &91.9\\
 &STR &1832 &458 &57.5 &56.6 &8 &8 &60.6 &60.8\\
\midrule
 \multirow{3}{*}{Vicuna-7B} &TQA &5000 &1800 &84.7 &85.4 &10 &10 &94.2 &93.8\\
 &NQ &5000 &1800 &87.6 &86.8 &10 &10 &96.6 &96.5\\
 &STR &1832 &458 &57.1 &57.1 &8 &8 &65.2 &68.1\\
 \midrule
 \multirow{3}{*}{Gemma-2B} &TQA &5000 &1800 &60.2 &56.2 &10 &10 &85.4 &82.8\\
 &NQ &5000 &1800 &87.9 &87.1 &10 &10 &96.7 &96.5\\
 &STR &1832 &458 &44.7 &44.5 &8 &8 &48.3 &47.9\\
  \midrule
 \multirow{2}{*}{Llama3.1-I-8B} &TQA &5000 &1800 &32.6 &25.7 &10 &10 &57.3 &52.8\\
 &STR &1832 &458 &26.9 &26.9 &8 &8 &40.3 &38.9\\
\bottomrule
\end{tabular}
\end{table}

\subsection{Response Refusal Rate}\label{ap:resp-refusal}
Depending on the LLM, and particularly with instruction fine-tuned models, the LLM may sometimes refuse to respond to queries, providing an "I don't know" type response instead. In our experiments, we are concerned with surfacing a factually correct response, when one exists, and therefore model the problem as a binary classification task of non-hallucination-vs-all, treating both true hallucinations as well as refusal responses under the same label. 
In order to validate that the hallucination label category is not dominated by refusal responses, in table \ref{tab:resp-refusal}, we analyse the \% responses containing any of the following common refusal phrases - ["don't know", "do not know", "don't have", "do not have", "can't", "cannot", "unable"]. We find that this is only a small proportion of responses and further manual inspection in fact indicates that the numbers reported are slight over-estimations since the phrases are also used in non refusal responses such as "\textit{Q: Who is featured on Puff Daddy's Can't Hold Me Down? A: 
jimmy page is featured on puff daddy's \underline{can't} hold me down.}" For Llama-3.1-Instruct 8B, being much more capable at answering STR (see \% Hallucinations in table \ref{tab:dataset-statistics}), the over-estimation is higher since the chain-of-thought reasoning often contains these phrases, eg. \textit{"Q: Do you have to pass through circle of lust to find Saladin in Dante's Inferno? A: dante's inferno is in three main circles: lust, gluttony, and the rest. saladin is mentioned in limbo. limbo is not one of the main circles. so the answer is no. in fact, it seems you have to pass through lust to get away from saladin. however, this is not an explicitly clear path in dante's inferno. it seems more likely that you simply \underline{cannot} pass through lust to find saladin."}. 

\begin{table}[h]
\centering
\caption{Refusal rate of LLMs. Shows \% of responses containing any one of a list of common refusal phrases.}\label{tab:resp-refusal}
\begin{tabular}{@{}llrrr@{}}
\toprule
 Model &Dataset &\multicolumn{3}{c}{\% Responses with a refusal phrase}\\
 & & Train: Greedy + Sampled & Test: Greedy &Test: Sampled \\
\midrule
\multirow{3}{*}{Llama-7B} &TQA &3.6 &0.1 &3.4 \\
 &NQ &4.5 &0.0 &5.2 \\
 &STR &4.4 &5.0 &3.9 \\
\midrule
 \multirow{3}{*}{Alpaca-7B} &TQA &0.8 &0.3 &0.9 \\
 &NQ &1.2 &0.9 &1.2 \\
 &STR &6.6 &5.7 &4.0 \\
\midrule
 \multirow{3}{*}{Vicuna-7B} &TQA &7.7 &0.9 &7.2 \\
 &NQ &9.6 &1.4 &10.0 \\
 &STR &8.1 &8.3 &6.1 \\
 \midrule
 \multirow{3}{*}{Gemma-2B} &TQA &0.9 &0.1 &1.1 \\
 &NQ &1.9 &0.0 &1.8 \\
 &STR &7.6 &3.7 &6.1 \\
  \midrule
 \multirow{2}{*}{Llama3.1-I-8B} &TQA &1.1 &0.2 &1.0 \\
 &STR &37.7 &22.1 &32.2 \\
\bottomrule
\end{tabular}
\end{table}

\subsection{Implementation Details}\label{ap:imp-details}
All probes are trained with a batch size of 128, using AdamW optimiser with linear warm-up for 5 epochs and cosine annealing for a maximum of 50 epochs. For each dataset and method, learning rate is selected from a coarse grid search $\in$ [0.5, 0.05, 0.005, 0.0005, 0.00005] using a held-out validation set.

\section{Additional Results}
\subsection{Hallucination Detection: Expanded Results for Table \ref{tab:main-average}}\label{ap:main-results-expanded}
Tables \ref{tab:main-greedytest} and \ref{tab:main-sampledtest} report the dataset-wise hallucination detection performance of CLAP against the uncertainty and activation probing baselines on greedy and sampled responses to prompts, respectively.

\begin{table*}[ht]
\centering
\caption{Hallucination detection performance of CLAP versus baselines on greedy responses to prompts, measured in AUC
scores, best is indicated in bold, second best is underlined. '-g' and '-s' denote the use of greedy responses only and the use of both greedy and sampled responses for training, respectively.}\label{tab:main-greedytest}
\begin{tabular}{@{}crrrrrrrrr@{}}
\toprule
Data &PE &AH-g &LP-g &NLP-g &CLAP-g &AH-s &LP-s &NLP-s &CLAP-s\\
\midrule
\multicolumn{10}{c}{Llama-7B}\\
\midrule
TQA &76.0 & 67.5 (0.1) &\textbf{82.0 (0.1)} &\underline{81.6 (0.4)} &\textbf{82.0 (0.0)} &72.1 (0.0) &80.1 (0.2) &81.2 (0.4) &\underline{81.6 (0.2)}\\
NQ & 80.1 & 75.0 (0.1) &85.9 (0.4) &\underline{86.2 (0.2)} &\textbf{86.6 (0.3)} &80.2 (0.7) &85.0 (0.2) &85.9 (0.1) &86.1 (0.4)\\
STR &50.9 &49.5 (1.1) &65.0 (1.3) &63.9 (0.8) &\textbf{65.7 (1.0)} &55.4 (0.6) &61.0 (1.1) &59.9 (2.3) &\underline{65.6 (0.6)}\\
Avg &69.0 &64.0 &77.7 &77.2 &\textbf{78.1} &69.2 &75.3 &75.6 &\underline{77.8}\\
\midrule
\multicolumn{10}{c}{Alpaca-7B}\\
\midrule
TQA &80.6 &82.8 (0.1) &87.6 (0.3) &\underline{87.7 (0.2)} &87.3 (0.7) &85.0 (0.0) &\textbf{89.3 (0.2)} &\textbf{89.3 (0.3)} &\textbf{89.3 (0.4)}\\
NQ &82.8 &83.7 (0.1) &88.0 (0.2) &88.1 (0.1) &88.2 (0.5) &84.2 (0.1) &\underline{88.5 (0.1)} &87.9 (0.2) &\textbf{89.1 (0.5)}\\
STR &72.3 & 77.2 (0.2) &80.1 (1.0) & 81.8 (0.2) & \textbf{84.2 (0.9)} &81.7 (0.3) &81.2 (1.9) &82.0 (1.1) &\underline{83.6 (1.1)} \\
Avg &78.6 &81.2 &85.2 &85.8 &\underline{86.6} &83.6 &86.3 &86.4 &\textbf{87.3}\\
\midrule
\multicolumn{10}{c}{Vicuna-7B}\\
\midrule
TQA & 81.4 & 81.5 (0.3) &\textbf{94.0 (0.2)} &\underline{93.8 (0.2)} &93.3 (0.3) &89.4 (1.1) &93.4 (0.5) &93.3 (0.4) &92.1 (1.4)\\
NQ &82.6 &76.2 (0.2) &88.1 (0.0) & 87.8 (0.2) &87.9 (0.2) &85.9 (0.6) &88.8 (0.7) &\underline{89.5 (0.2)} &\textbf{89.6 (0.5)}\\
STR & 77.1 & 76.2 (0.3) &83.1 (0.3) &83.1 (0.5) &\underline{84.2 (0.2)} &81.8 (0.2) &\underline{84.2 (0.8)} &83.7 (0.7) &\textbf{84.3 (0.7)}\\
Avg &80.4 &78.0 &88.4 &88.2 &88.5 &85.7 &\textbf{88.8} &\textbf{88.8} &\underline{88.7}\\
\midrule
\multicolumn{10}{c}{Gemma-2B}\\
\midrule
TQA &70.6 &\underline{80.4 (0.7)} &75.1 (0.4) &75.6 (0.9) &78.0 (0.4) &79.0 (0.0) &77.1 (0.2) &77.8 (0.5) &\textbf{80.6 (0.3)}\\
NQ &63.7 &\textbf{84.4 (0.5)} &78.7 (0.6) &80.3 (0.9) &82.1 (1.5) &\underline{83.5 (0.1)} &80.2 (0.8) &80.8 (0.4) &83.1 (0.1)\\
STR &54.0 &55.7 (1.6) &56.8 (0.9) &57.6 (1.2) &\underline{58.1 (0.7)} &\textbf{59.6 (2.0)} &52.9 (4.5) &56.3 (2.6) &54.4 (3.8)\\
Avg &62.8 &\underline{73.5} &70.2 &71.2 &72.7 &\textbf{74.1} &70.1 &71.6 &72.7 \\
\midrule
\multicolumn{10}{c}{Llama3.1-Instruct-8B}\\
\midrule
TQA &83.3 &81.2 (1.6) &83.3 (0.6) &83.9 (0.6) &\underline{86.3 (0.1)} &85.0 (0.1) &85.3 (0.5) &86.1 (0.6) &\textbf{88.3 (0.1)}\\
STR &\textbf{56.0} &\underline{51.9 (1.9)} &48.9 (1.0) &49.2 (1.1) &50.0 (0.6) &50.3 (2.2) &50.3 (1.7) &50.5 (1.2) &49.7 (2.5)\\
Avg &\textbf{69.7} &66.6 &66.1 &66.5 &68.1 &67.6 &67.8 &68.3 &\underline{69.0}\\
\bottomrule
\end{tabular}
\end{table*}

\begin{table*}[h]
\centering
\caption{Hallucination detection performance of CLAP versus baselines on sampled responses to prompts, measured in AUC
scores, best is indicated in bold, second best is underlined. '-g' and '-s' denote the use of greedy responses only and the use of both greedy and sampled responses for training, respectively.}\label{tab:main-sampledtest}
\begin{tabular}{@{}crrrrrrrrr@{}}
\toprule
Data &PE &AH-g &LP-g &NLP-g &CLAP-g &AH-s &LP-s &NLP-s &CLAP-s \\
\midrule
\multicolumn{10}{c}{Llama-7B}\\
\midrule
TQA &88.0 &81.0 (0.5) &82.5 (0.5) &87.0 (0.7) &85.8 (1.8) &86.9 (0.1) &90.0 (0.7) &\underline{90.5 (0.0)} &\textbf{91.2 (0.1)}\\
NQ &90.3 &82.0 (0.3) &79.7 (2.9) &88.1 (0.3) &85.3 (0.6) &90.3 (0.2) &90.5 (0.8) &\underline{91.2 (0.7)} &\textbf{92.3 (0.2)} \\
STR &74.6 &71.5 (3.4) &65.2 (3.8) &57.4 (8.2) &51.6 (3.8) &82.1 (0.1) &\underline{85.8 (0.1)} &85.3 (0.1) &\textbf{86.4 (0.3)} \\
Avg &84.3 &78.2 &75.8 &77.5 &74.3 &86.5 &88.8 &\underline{89.0} &\textbf{89.9} \\
\midrule
\multicolumn{10}{c}{Alpaca-7B}\\
\midrule
TQA &84.9 &84.2 (0.3) &87.5 (0.2) &87.2 (0.2) &87.4 (0.4)  &86.7 (0.0) &\underline{88.8 (1.0)} &88.6 (0.1) &\textbf{90.3 (0.1)} \\
NQ &85.6 &81.4 (0.0) &84.3 (0.2) &83.4 (0.2) &83.1 (1.6) &83.2 (0.1) &85.1 (1.9) &\underline{87.5 (0.2)} &\textbf{88.3 (0.6)}  \\
STR &68.5 &79.0 (0.3) &82.6 (0.1) &82.5 (0.1) &83.8 (0.2) &81.6 (0.2) &84.2 (0.5) &\underline{84.2 (0.4)} &\textbf{85.6 (0.3)}  \\
Avg &79.7 &81.5 &84.8 &84.4 &84.8 &83.8 &86.0 &\underline{86.8} &\textbf{88.1} \\
\midrule
\multicolumn{10}{c}{Vicuna-7B}\\
\midrule
TQA &92.4 &87.4 (1.9) &92.5 (0.9) &92.7 (0.9) &92.1 (1.1) &93.4 (0.6) &\textbf{95.4 (0.3)} &\underline{95.3 (0.6)} &95.2 (0.4)  \\
NQ &93.8 &89.2 (0.2) &91.3 (0.9) &91.4 (0.9) &92.1 (0.9) &92.9 (0.4) &94.9 (0.5) &\textbf{95.4 (0.1)} &\underline{95.2 (0.4)}  \\
STR &77.4 &84.3 (0.4) &88.6 (0.4) &88.6 (0.1) &\underline{89.1 (0.1)} &86.3 (0.1) &\underline{89.1 (0.2)} &\underline{89.1 (0.1)} &\textbf{89.5 (0.1)} \\
Avg &87.9 &87.0 &90.8 &90.9 &91.1 &90.8 &\underline{93.1} &\textbf{93.3} &\textbf{93.3}  \\
\midrule
\multicolumn{10}{c}{Gemma-2B}\\
\midrule
TQA &82.1 &77.4 (1.7) &75.8 (1.1) &75.8 (2.8) &81.0 (0.8) &\underline{87.3 (0.1)} &85.8 (0.0) &86.1 (1.1) &\textbf{88.8 (0.0)}  \\
NQ &81.7 &81.4 (2.4) &70.2 (1.8) &73.6 (2.4) &78.3 (3.6) &\textbf{88.8 (0.1)} &82.1 (0.6) &84.6 (1.0) &\underline{87.4 (0.9)}  \\
STR &\underline{53.4} &49.0 (0.3) &50.3 (0.5) &49.3 (0.1) &49.6 (0.4) &\textbf{54.3 (0.8)} &52.3 (0.7) &52.8 (0.5) &53.2 (0.3)  \\
Avg &72.4 &69.3 &65.4 &66.2 &69.7 &\textbf{76.8} &73.4 &74.5 &\underline{76.5} \\
\midrule
\multicolumn{10}{c}{Llama3.1-Instruct-8B}\\
\midrule
TQA &87.8 &84.3 (0.3) &86.8 (0.2) &87.1 (0.3) &88.6 (0.1) &89.4 (0.0) &90.2 (0.0) &\underline{90.4 (0.3)} &\textbf{91.8 (0.2)}  \\
STR &\textbf{59.7} &50.8 (0.5) &49.8 (0.4) &49.0 (0.9) &50.7 (0.5) &50.2 (1.7) &56.9 (0.6) &57.8 (0.5) &\underline{58.1 (0.7)}  \\
Avg &73.8 &67.6 &68.3 &68.0 &69.6 &69.8 &73.5 &\underline{74.1} &\textbf{74.9}\\
\bottomrule
\end{tabular}
\end{table*}

\subsection{Analysis of Train-time Batching Strategy of Sampled Responses}\label{ap:batching-sampled-responses}
In table \ref{tab:batching-strategy}, we compare the effect of arranging sampled responses of the same prompt to be in the same training batch, denoted as \textit{prompt-wise sampling (pw)}, against the strategy of randomly sampling each batch from the set of all sampled responses across prompts, denoted as \textit{random sampling (rs)}. On greedy responses, we observe minor improvements using the \textit{prompt-wise sampling} strategy for each method. On sampled responses, we observe significant gains using the \textit{prompt-wise sampling} strategy. For the experiments reported in the main-text we use the \textit{prompt-wise sampling} strategy when training on sampled responses.

\begin{table*}[h]
\centering
\caption{Comparing effect of train-time batching strategy on hallucination detection performance, measured in AUC
scores, best is indicated in bold, second best is underlined.}\label{tab:batching-strategy}
\begin{tabular}{@{}ccrrrrrr@{}}
\toprule
Model &Data &LP-rs &NLP-rs &CLAP-rs &LP-pw &NLP-pw &CLAP-pw\\
\midrule
\multicolumn{8}{c}{Greedy Test Responses}\\
\midrule
L-7B &TQA &79.9 (1.0) &80.1 (0.2) &80.8 (0.9) &80.1 (0.2) &81.2 (0.4) &\underline{81.6 (0.2)}\\
L-7B &NQ &84.3 (0.7) &85.0 (0.4) &84.7 (0.8) &85.0 (0.2) &85.9 (0.1) &86.1 (0.4)\\
L-7B &STR &60.6 (1.2) &59.9 (3.3) &66.5 (1.8) &61.0 (1.1) &59.9 (2.3) &65.6 (0.6)\\
&Avg &74.9 &75.0 &\underline{77.3} &75.3 &75.6 &\textbf{77.8}\\
\midrule
A-7B &TQA &89.4 (0.1) &88.9 (0.4) &89.1 (0.4) &89.3 (0.2) &89.3 (0.3) &89.3 (0.4)\\
A-7B &NQ &87.8 (0.1) &87.2 (0.3) &89.5 (0.5) &88.5 (0.1) &87.9 (0.2) &89.1 (0.5)\\
A-7B &STR &81.0 (1.5) &82.3 (1.7) &83.0 (0.3) &81.2 (1.9) &82.0 (1.1) &83.6 (1.1) \\
&Avg &86.1 &86.1 &\underline{87.2} &86.3 &86.4 &\textbf{87.3}\\
\midrule
V-7B &TQA &92.7 (0.8) &92.9 (0.5) &92.6 (0.5) &93.4 (0.5) &93.3 (0.4) &92.1 (1.4)\\
V-7B &NQ &89.6 (0.2) &89.5 (0.6) &89.1 (0.4) &88.8 (0.7) &89.5 (0.2) &89.6 (0.5)\\
V-7B &STR &84.1 (0.7) &83.6 (0.4) &83.4 (0.8) &84.2 (0.8) &83.7 (0.7) &84.3 (0.7)\\
&Avg &\textbf{88.8} &88.6 &88.4 &\textbf{88.8} &\textbf{88.8} &\underline{88.7}\\
\midrule
G-2B &TQA &76.6 (0.2) &76.8 (0.9) &79.9 (0.9) &77.1 (0.2) &77.8 (0.5) &80.6 (0.3)\\
G-2B &NQ &79.9 (0.4) &81.0 (0.2) &82.0 (0.7) &80.2 (0.8) &80.8 (0.4) &83.1 (0.1)\\
G-2B &STR &55.2 (1.7) &58.5 (0.0) &56.3 (4.3) &52.9 (4.5) &56.3 (2.6) &54.4 (3.8)\\
&Avg &70.6 &\underline{72.1} &\textbf{72.7} &70.1 &71.6 &\textbf{72.7} \\
\midrule
L3I-8B &TQA &84.7 (0.7) &85.1 (0.7) &86.9 (0.7) &85.3 (0.5) &86.1 (0.6) &88.3 (0.1)\\
L3I-8B &STR &49.9 (1.0) &49.3 (0.7) &48.9 (1.9) &50.3 (1.7) &50.5 (1.2) &49.7 (2.5)\\
&Avg &67.3 &67.2 &67.9 &67.8 &\underline{68.3} &\textbf{69.0}\\
\midrule
\multicolumn{8}{c}{Sampled Test Responses}\\
\midrule
L-7B &TQA &74.0 (1.2) &77.4 (3.1) &81.3 (1.4) &90.0 (0.7) &90.5 (0.0) &91.2 (0.1)\\
L-7B &NQ &72.7 (0.8) &73.8 (1.4) &76.2 (2.7) &90.5 (0.8) &91.2 (0.7) &92.3 (0.2) \\
L-7B &STR &85.7 (0.2) &85.6 (0.1) &86.5 (0.3) &85.8 (0.1) &85.3 (0.1) &86.4 (0.3) \\
&Avg &77.5 &78.9 &81.3 &88.8 &\underline{89.0} &\textbf{89.9} \\
\midrule
A-7B &TQA &77.1 (0.8) &77.2 (3.4) &75.3 (3.6) &88.8 (1.0) &88.6 (0.1) &90.3 (0.1) \\
A-7B &NQ &71.6 (1.2) &69.3 (3.7) &68.6 (0.4) &85.1 (1.9) &87.5 (0.2) &88.3 (0.6)  \\
A-7B &STR &84.4 (0.6) &84.6 (0.6) &85.4 (0.3) &84.2 (0.5) &84.2 (0.4) &85.6 (0.3)  \\
&Avg &77.7 &77.0 &76.5 &86.0 &\underline{86.8} &\textbf{88.1} \\
\midrule
V-7B &TQA &77.0 (6.3) &75.4 (2.4) &73.1 (3.9) &95.4 (0.3) &95.3 (0.6) &95.2 (0.4)  \\
V-7B &NQ &76.5 (5.4) &76.7 (7.3) &79.3 (2.8) &94.9 (0.5) &95.4 (0.1) &95.2 (0.4)  \\
V-7B &STR &89.3 (0.2) &89.1 (0.2) &88.9 (0.4) &89.1 (0.2) &89.1 (0.1) &89.5 (0.1) \\
&Avg &80.9 &80.4 &80.4 &\underline{93.1} &\textbf{93.3} &\textbf{93.3}  \\
\midrule
G-2B &TQA &85.3 (0.3) &85.7 (0.5) &88.3 (0.2) &85.8 (0.0) &86.1 (1.1) &88.8 (0.0)  \\
G-2B &NQ &83.2 (0.4) &84.9 (0.5) &87.5 (0.7) &82.1 (0.6) &84.6 (1.0) &87.4 (0.9)  \\
G-2B &STR &52.7 (0.4) &53.4 (0.5) &53.3 (0.5) &52.3 (0.7) &52.8 (0.5) &53.2 (0.3)  \\
&Avg &73.7 &74.7 &\underline{76.3} &73.4 &74.5 &\textbf{76.5} \\
\midrule
L3I-8B &TQA &90.4 (0.1) &90.4 (0.3) &91.4 (0.2) &90.2 (0.0) &90.4 (0.3) &91.8 (0.2) \\
L3I-8B &STR &57.5 (0.6) &57.3 (0.4) &57.3 (0.3) &56.9 (0.6) &57.8 (0.5) &58.1 (0.7)  \\
&Avg &73.9 &73.8 &\underline{74.4} &73.5 &74.1 &\textbf{74.9}\\
\bottomrule
\end{tabular}
\end{table*}

\subsection{Hallucination Mitigation: CLAP with Random Sampling}\label{ap:mitigation-random-sampling}
Table \ref{tab:mitigation-clap-with-random-sampling} shows the results of combining fine-grained detection using CLAP with random sampling. We see that even though random sampling (Alt) has very low non-hallucination rate, combining Def and Alt using +CLAP-I can still often result in improvements over default decoding (except for Llama 7B). Using +CLAP-II results in significantly lower abstention rate compared to Def+Abs, while again maintaining high non-hallucination rate among non-abstained responses.

\begin{table*}[h]
\centering
\caption{Mitigating hallucinations with fine-grained detection using CLAP applied to Random Sampling. 
Second block shows the \% of non-hallucinated responses for each of the five response generation strategies. Third and forth blocks show the \% of responses abstained and the \% of responses that were abstained but were non-hallucinated (NH), respectively. For results with +CLAP-I and +CLAP-II we report mean across three seeds. * denotes \%non-hallucinations among non-abstained responses.}\label{tab:mitigation-clap-with-random-sampling}
\begin{NiceTabular}{@{}c|rrrrr|rr|rr@{}}
\toprule
&\multicolumn{5}{c}{\% Non-Hallucinations $\uparrow$} &\multicolumn{2}{c}{\%Abs $\downarrow$} &\multicolumn{2}{c}{\%Abs but NH $\downarrow$}\\
Data &Def &Def+Abs\textsuperscript{*} &Alt &+CLAP-I &+CLAP-II\textsuperscript{*} &Def+Abs &+CLAP-II &Def+Abs &+CLAP-II\\
\midrule
L-7B TQA &57.3 &75.6 &22.8 &51.8 &71.8 &36.0 &29.7 &9.1 &6.9\\
L-7B STR &60.0 &69.9 &32.3 &51.9 &59.9 &45.0 &22.7 &21.4 &10.9 \\
\midrule
A-7B TQA &34.2 &68.3 &26.9 &37.6 &67.3 &62.7 &50.4 &8.7 &6.3\\
A-7B STR &43.4 &67.0 &39.7 &48.2 &61.4 &44.3 &31.6 &6.1 &3.0\\
\midrule
V-7B TQA &14.6 &59.2 &6.6 &15.6 &53.7 &80.7 &72.6 &3.2 &2.8 \\
V-7B STR &42.8 &65.6 &32.8 &51.2 &59.9 &40.3 &22.6 &3.6 &2.1 \\
\midrule
Average &42.1 &67.6 &26.9 &42.7 &62.3 &51.5 &38.3 &8.7 &5.3\\
\bottomrule
\end{NiceTabular}
\end{table*}

\subsection{Hallucination Mitigation: CLAP versus Last Layer Probing}\label{ap:mitigation-clap-vs-baselines}
Table \ref{tab:mitigation-CLAPI-vs-baselines} compares mitigation using +CLAP-I against using baseline probes. We see that +CLAP-I results in better overall non-hallucination rates compared to the two baselines and that this stems from the higher H->NH replacements using +CLAP-I. Tables \ref{tab:mitigation-CLAPII-vs-baselines} and \ref{tab:mitigation-changes-CLAPII-vs-baselines} compare mitigation using +CLAP-II against using baseline probes. We see that +CLAP-II results in better overall non-hallucination rates, while maintaining comparable abstention rates, and that again the improvement stems from the higher H->NH replacements using +CLAP-II.

\begin{table*}[h]
\centering
\caption{Hallucination Mitigation: CLAP-I versus Last Layer Probing when combined with DoLa, best is indicated in bold. Second block shows the \% of non-hallucinated responses using each method. Third and fourth blocks show the ratio of \% H->NH (or NH->H) replacements using LP-I/NLP-I/CLAP-I against \% H->NH (or NH->H) replacements using Alt. Results averaged across TQA and STR, across three seeds.}\label{tab:mitigation-CLAPI-vs-baselines}
\begin{NiceTabular}{@{}c|rrr|rrr|rrr@{}}
\toprule
&\multicolumn{3}{c}{\% Non-Hallucinations $\uparrow$} &\multicolumn{3}{c}{H -> NH $\uparrow$} &\multicolumn{3}{c}{NH -> H $\downarrow$} \\
LLM &+LP-I &+NLP-I &+CLAP-I &+LP-I &+NLP-I  &+CLAP-I &+LP-I &+NLP-I  &+CLAP-I\\
\midrule
Llama 7B &60.2 &60.0 &\textbf{61.1} &55.8 &45.8 &\textbf{57.6} &34.4 &\textbf{26.8} &32.8 \\
Alpaca 7B &51.6 &52.7 &\textbf{53.4} &77.2 &78.1 &\textbf{80.1} &26.4 &\textbf{21.6} &21.8 \\
\bottomrule
\end{NiceTabular}
\end{table*}

\begin{table*}[h]
\centering
\caption{Hallucination Mitigation: CLAP-II versus Last Layer Probing when combined with DoLa, best is indicated in bold. Second block shows the \% of non-hallucinated responses using each method. Third and fourth blocks show the \% of responses abstained and the \% of responses that were abstained but were non-hallucinated (NH), respectively. * denotes \%non-hallucinations among non-abstained responses. Results averaged across TQA and STR, across three seeds.}\label{tab:mitigation-CLAPII-vs-baselines}
\begin{NiceTabular}{@{}c|rrr|rrr|rrr@{}}
\toprule
&\multicolumn{3}{c}{\% Non-Hallucinations $\uparrow$} &\multicolumn{3}{c}{\%Abs $\downarrow$} &\multicolumn{3}{c}{\%Abs but NH $\downarrow$}\\
LLM &+LP-II\textsuperscript{*} &+NLP-II\textsuperscript{*} &+CLAP-II\textsuperscript{*}  &+LP-II &+NLP-II &+CLAP-II &+LP-II &+NLP-II &+CLAP-II\\
\midrule
Llama 7B &67.8 &68.5 &\textbf{69.5} &\textbf{16.4} &18.5 &16.8 &4.9 &5.3 &\textbf{4.7} \\
Alpaca 7B &66.6 &64.9 &\textbf{66.7} &33.1 &\textbf{29.2} &30.2 &5.6 &\textbf{4.2} &4.5 \\
\bottomrule
\end{NiceTabular}
\end{table*}

\begin{table*}[h]
\centering
\caption{Comparison of LP-II/NLP-II with DoLa versus CLAP-II with DoLa, best is indicated in bold. Shows the ratio of \% H->NH (or NH->H) replacements using LP-II/NLP-II/CLAP-II against \% H->NH (or NH->H) replacements using Alt. Results averageed across TQA and STR, across three seeds.}\label{tab:mitigation-changes-CLAPII-vs-baselines}
\begin{NiceTabular}{@{}c|rrr|rrr@{}}
\toprule
&\multicolumn{3}{c}{H -> NH $\uparrow$} &\multicolumn{3}{c}{NH -> H $\downarrow$}\\
LLM &+LP-II &+NLP-II  &+CLAP-II &+LP-II &+NLP-II  &+CLAP-II\\
\midrule
Llama 7B &41.0 &27.5 &\textbf{42.6} &19.8 &\textbf{11.7} &19.0 \\
Alpaca 7B &49.2 &57.0 &\textbf{60.4} &\textbf{3.8} &4.2 &4.2 \\
\bottomrule
\end{NiceTabular}
\end{table*}

\section{Design Ablations}
\subsection{Comparing CLAP with Token-wise Attention-pooling}\label{ap:clap-vs-att-pool}
In table \ref{tab:ablation-att-pool} we compare CLAP against attention pooling \cite{ch-wang-etal-2024-attpool}, which implements a learnable query vector followed by softmax pooling to aggregate token-wise activations at each layer before training a logistic regression probe on the pooled activation vector. Following the original work, we train 2L attention-pooling probes where L denotes the number of LLM decoder layers and probes are trained at both layer output as well as attention output (after residual connection) positions. After training the 2L probes, the individual probe weights are frozen and an ensemble logistic regression probe is trained on the output of the individual probes. \textbf{Att-Pool (MA)} denotes the best individual probe out of 2L probes (chosen using in-distribution validation data), while \textbf{Att-Pool-Ens} denotes the ensemble probe. We implement attention pooling with 20 tokens, taking either the last 20 or padding to 20 with zero vectors, as required\footnote{We train all probes including CLAP on 2000 samples instead of the 5000 samples used for the main experiments, due to the GPU memory constraint of loading token-wise activations for all layers when training.}.  We find that while token-wise attention pooling slightly outperforms CLAP on in-distribution testing, CLAP significantly outperforms in the out-of-distribution setting, demonstrating its superiority.

\begin{table}[h]
\centering
\caption{Comparing CLAP with token-wise attention-pooling, using AUC scores on test data, best is indicated in bold. Results on Gemma 2B, using greedy responses for train and test.}\label{tab:ablation-att-pool}
\begin{tabular}{@{}crr@{}}
\toprule
Probe &In Distribution &Out Of Distribution\\
 &TQA->TQA &TQA->City\\
\midrule
Att-Pool (MA) &\textbf{77.7 (1.2)} &72.3 (4.4) \\
Att-Pool-Ens &77.3(0.4) &66.4 (4.9) \\
CLAP &76.3 (0.3) &\textbf{79.0 (3.9)} \\
\bottomrule
\end{tabular}
\end{table}

\subsection{Analysis of Hyper-parameter Choices for CLAP}\label{ap:hyp-choices}
Table \ref{tab:ablation-design-auc} reports the effect of varying the two architectural hyper-parameters $n_{enc}$ and $d_{model}$ on the validation data for the Alpaca 7B, Vicuna 7B, Gemma 2B and Llama3.1-Instruct 8B models.

\begin{table}[h]
\centering
\caption{Analysis of hyper-parameter choices for CLAP, using AUC scores on validation data.}\label{tab:ablation-design-auc}
\begin{tabular}{@{}rr|rrr|rrr@{}}
\toprule
$n_{enc}$ &$d_{model}$ &TQA & NQ  & STR  &TQA & NQ  & STR \\
\midrule
 & &\multicolumn{3}{c}{Alpaca 7B} &\multicolumn{3}{c}{Vicuna 7B}\\
\midrule
1 &128 &87.3 (0.4) &89.0 (1.2) &83.8 (0.2) &93.0 (0.9) &88.1 (0.1) &85.1 (1.4) \\
1 &256 &87.2 (0.5) &\textbf{89.0 (1.0)} &83.6 (0.2)  &92.8 (0.8) &88.3 (0.3) &84.7 (1.2) \\
1 &512 &\textbf{87.4 (0.5)} &88.6 (1.2) &83.8 (0.6) &92.8 (0.9) &88.1 (0.3) &85.3 (1.3)  \\
1 &1024 &87.3 (0.5) &88.5 (1.0) &83.7 (0.3) &\textbf{93.0 (0.8)} &88.1 (0.2) &85.1 (1.4)   \\
1 &2048 &\textbf{87.4 (0.5)} &88.3 (1.0) &83.5 (0.3) &92.8 (0.9) &\textbf{88.3 (0.2)} &85.4 (1.4)   \\
1 &4096* &87.3 (0.7) &88.6 (1.0) &83.4 (0.4) &92.9 (0.8) &88.1 (0.1) &\textbf{85.6 (1.6)}   \\
2 &128 &87.2 (0.7) &88.9 (1.3) &83.4 (0.3) &92.9 (0.8) &87.9 (0.1) &84.8 (1.5)   \\
2 &256 &87.3 (0.5) &\textbf{89.0 (1.0)} &83.4 (0.3) &92.7 (0.8) &88.2 (0.2) &84.8 (1.2)   \\
2 &512 &87.3 (0.4) &88.7 (1.2) &83.7 (0.5) &92.8 (0.8) &88.2 (0.3) &85.0 (1.0)   \\
2 &1024 &87.2 (0.5) &88.7 (1.2) &83.8 (0.4) &\textbf{93.0 (0.8)} &88.0 (0.1) &85.0 (1.6)   \\
2 &2048 &87.2 (0.4) &88.6 (1.3) &\textbf{83.9 (0.4)} &93.0 (0.9) &88.2 (0.3) &85.2 (1.6)   \\
2 &4096* &87.0 (0.4) &88.3 (1.3) &83.7 (0.8) &92.6 (0.8) &88.1 (0.3) &85.4 (1.5)   \\
\midrule
 & &\multicolumn{3}{c}{Gemma 2B} &\multicolumn{3}{c}{Llama3.1-Instruct 8B}\\
\midrule
1 &128 &80.8 (0.6) &79.5 (0.8) &59.1 (1.6) &87.6 (0.5) &- &55.5 (0.6) \\
1 &256 &80.8 (0.5) &79.3 (0.4) &59.2 (2.3) &87.9 (0.6) &- &55.9 (0.7) \\
1 &512 &80.9 (0.3) &79.4 (0.5) &59.5 (1.3) &87.9 (0.4) &- &55.5 (0.3) \\
1 &1024 &80.6 (0.6) &\textbf{79.6 (0.1)} &\textbf{60.0 (1.9)} &87.8 (0.4) &- &55.9 (0.3)  \\
1 &2048 &\textbf{81.3 (0.4)}* &79.1 (1.0)* &59.6 (2.6)* &87.8 (0.4) &- &56.0 (0.3)  \\
1 &4096* &- &- &-  &87.2 (0.4) &- &55.5 (0.6)  \\
2 &128 &80.7 (0.8) &79.5 (0.2) &59.5 (2.0) &87.9 (0.6) &- &55.7 (0.4) \\
2 &256 &80.8 (0.7) &\textbf{79.6 (0.1)} &59.3 (2.0) &88.2 (0.6) &- &55.8 (1.0)  \\
2 &512 &80.6 (0.6) &79.3 (0.4) &59.2 (1.4) &88.2 (0.3) &- &56.1 (0.6)  \\
2 &1024 &80.6 (0.9) &79.1 (0.4) &58.8 (1.8) &\textbf{88.4 (0.3)} &- &55.7 (0.9) \\
2 &2048 &80.7 (0.5)* &78.9 (0.8)* &58.0 (1.9)* &88.3 (0.2) &- &55.8 (0.6) \\
2 &4096* &- &- &- &87.8 (0.3) &- &\textbf{56.7 (0.7)}  \\
\bottomrule
\end{tabular}
\end{table}

\end{document}